\documentclass[conference]{IEEEtran}
\IEEEoverridecommandlockouts

\usepackage{cite}
\usepackage{enumitem}
\usepackage{amsmath,amssymb,amsfonts}
\usepackage{algorithmic}
\usepackage{graphicx}
\usepackage{textcomp}
\usepackage{xcolor}
\usepackage{forest}
\usetikzlibrary{trees,positioning,shapes,shadows,arrows.meta}
\usepackage{tikz}

\makeatletter
\newcommand{\linebreakand}{%
  \end{@IEEEauthorhalign}
  \hfill\mbox{}\par
  \mbox{}\hfill\begin{@IEEEauthorhalign}
}
\makeatother

\begin{document}

\title{From Pixels to Prose: Advancing Multi-Modal Language Models for Remote Sensing}

\author{
    \IEEEauthorblockN{
        Xintian Sun\textsuperscript{1, 3},
        Benji Peng\textsuperscript{*, 2, 6},
        Charles Zhang\textsuperscript{2, 6},
        Fei Jin\textsuperscript{4}, 
        Qian Niu\textsuperscript{5}, 
        Junyu Liu\textsuperscript{5}, \\
        Keyu Chen\textsuperscript{6},
        Ming Li\textsuperscript{6},
        Pohsun Feng\textsuperscript{7},
        Ziqian Bi\textsuperscript{8},
        Ming Liu\textsuperscript{8},
        Xinyuan Song\textsuperscript{9},
        Yichao Zhang\textsuperscript{10}
    }
    \IEEEauthorblockA{
        \textsuperscript{1}Simon Fraser University, Canada
    }
    \IEEEauthorblockA{
        \textsuperscript{2}AppCubic, USA
    }
    \IEEEauthorblockA{
        \textsuperscript{3}University of Minnesota - Twin Cities, USA
    }
    \IEEEauthorblockA{
        \textsuperscript{4}Depth LLC, USA
    }
    \IEEEauthorblockA{
        \textsuperscript{5}Kyoto University, Japan
    }
    \IEEEauthorblockA{
        \textsuperscript{6}Georgia Institute of Technology, USA
    }
    \IEEEauthorblockA{
        \textsuperscript{7}National Taiwan Normal University, ROC
    }
    \IEEEauthorblockA{
        \textsuperscript{8}Purdue University, USA
    }
    \IEEEauthorblockA{
        \textsuperscript{9}Emory University, USA
    }
    \IEEEauthorblockA{
        \textsuperscript{10}The University of Texas at Dallas, USA
    }
    \IEEEauthorblockA{
        *Corresponding Email: benji@appcubic.com
    }
}

\maketitle


\begin{abstract}
Remote sensing has evolved from simple image acquisition to complex systems capable of integrating and processing visual and textual data. This review examines the development and application of multi-modal language models (MLLMs) in remote sensing, focusing on their ability to interpret and describe satellite imagery using natural language. We cover the technical underpinnings of MLLMs, including dual-encoder architectures, Transformer models, self-supervised and contrastive learning, and cross-modal integration. The unique challenges of remote sensing data--varying spatial resolutions, spectral richness, and temporal changes--are analyzed for their impact on MLLM performance. Key applications such as scene description, object detection, change detection, text-to-image retrieval, image-to-text generation, and visual question answering are discussed to demonstrate their relevance in environmental monitoring, urban planning, and disaster response. We review significant datasets and resources supporting the training and evaluation of these models. Challenges related to computational demands, scalability, data quality, and domain adaptation are highlighted. We conclude by proposing future research directions and technological advancements to further enhance MLLM utility in remote sensing.
\end{abstract}


\section{Introduction}

Remote sensing has evolved from basic image capture to advanced systems that integrate visual and textual data processing. This progression has led to the development of multi-modal language models (MLLMs) capable of interpreting and describing satellite imagery using natural language \cite{zhang2024earthgpt, zhan2024skyeyegpt}. Such integration enhances the analysis of Earth observation data, making it more accessible and informative.

Applying MLLMs to remote sensing data has significantly improved automated Earth observation analysis. These models support vital applications in environmental monitoring, urban planning, and disaster response by enabling more efficient and accurate extraction of information from satellite imagery \cite{guo2024skysense, mi2024knowledge}. This advancement enhances decision-making processes across various sectors.

This review examines the integration of MLLMs in remote sensing, focusing on their technical foundations, current applications, and potential future developments. The objective is to provide a comprehensive overview of how these models enhance the interpretation of satellite imagery and to identify areas for further research and application.

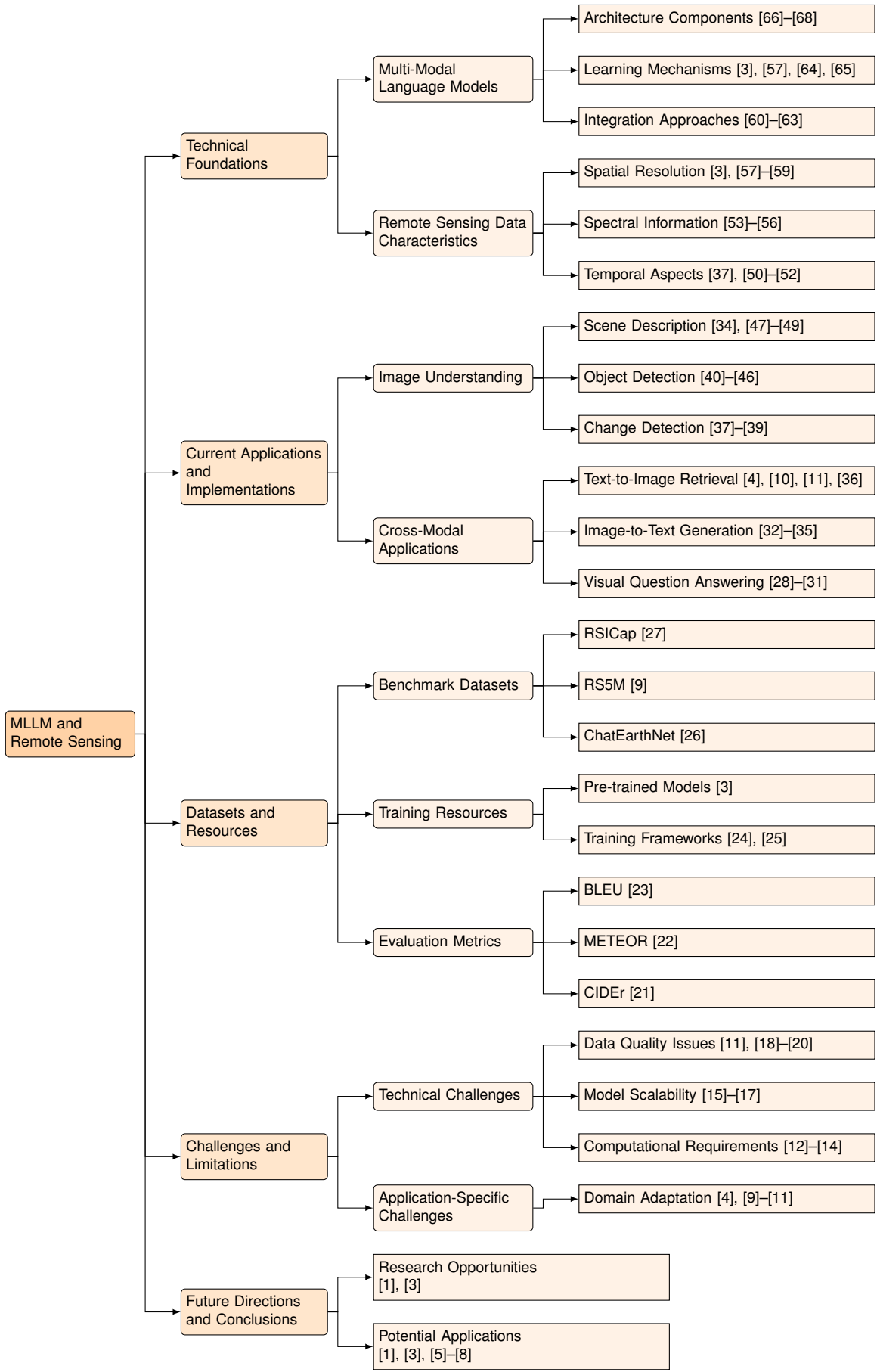
\begin{figure*}
    \centering
    
\tikzset{
    basic/.style  = {draw, align=left, font=\sffamily, rectangle},
    root/.style   = {basic, rounded corners=2pt, thin, fill=orange!35, text width=2.8cm},
    level1/.style = {basic, thin, rounded corners=2pt, fill=orange!20, text width=3.2cm},
    level2/.style = {basic, thin, rounded corners=2pt, fill=orange!10, text width=3.5cm},
    leaf/.style   = {basic, thin, fill=orange!10, text width=6.7cm},
    edge from parent/.style={draw=black, edge from parent fork right},
    level distance=1.2cm,
}

\begin{forest}
for tree={
    grow=east,
    scale=0.75, 
    growth parent anchor=west,
    parent anchor=east,
    child anchor=west,
    l sep=8mm, 
    s sep=4mm,  
    edge path={
        \noexpand\path[\forestoption{edge},->, >={latex}] 
             (!u.parent anchor) -- +(5pt,0pt) |- (.child anchor)
             \forestoption{edge label};
    },
    align=left, 
}
[MLLM and \\ Remote Sensing, root, 
    [Future Directions \\ and Conclusions, level1,
        [Potential Applications \\ \cite{guo2024skysense, yuan2024surveillance, zhang2024earthgpt, zheng2023marinegpt, peng2024securing, peng2024jailbreaking}, leaf]
        [Research Opportunities \\ \cite{guo2024skysense, zhang2024earthgpt}, leaf]
    ]
    [Challenges and \\ Limitations, level1,
        [Application-Specific \\ Challenges, level2,
            [Domain Adaptation \cite{zhang2023rs5m, mi2024knowledge, pan2024pir, mikriukov2022deep}, leaf]
        ]
        [Technical Challenges, level2,
            [Computational Requirements \cite{basu2024understanding, hadji2023context, gopalkrishnan2024multi}, leaf]
            [Model Scalability \cite{aghajanyan2023scaling, tian2022resformer, liu2024sphinx}, leaf]
            [Data Quality Issues \cite{pasquarella2023comprehensive, thai2022riesz, song2020unsupervised, mikriukov2022deep}, leaf]
        ]
    ]
    [Datasets and \\ Resources, level1,
        [Evaluation Metrics, level2,
            [CIDEr \cite{vedantam2015cider}, leaf]
            [METEOR \cite{banerjee2005meteor}, leaf]
            [BLEU \cite{papineni2002bleu}, leaf]
        ]
        [Training Resources, level2,
            [Training Frameworks \cite{huang2024language, berg2023joint}, leaf]
            [Pre-trained Models \cite{guo2024skysense}, leaf]
        ]
        [Benchmark Datasets, level2,
            [ChatEarthNet \cite{yuan2024chatearthnet}, leaf]
            [RS5M \cite{zhang2023rs5m}, leaf]
            [RSICap \cite{hu2023rsgpt}, leaf]
        ]
    ]
    [Current Applications \\ and \\ Implementations, level1,
        [Cross-Modal \\ Applications, level2,
            [Visual Question Answering \cite{lobry2020rsvqa, zheng2021mutual, siebert2022multi, hackel2023lit}, leaf]
            [Image-to-Text Generation \cite{shen2020remote, li2020multi, silva2024large, zhan2023rsvg}, leaf]
            [Text-to-Image Retrieval \cite{yuan2022remote, mi2024knowledge, pan2024pir, mikriukov2022deep}, leaf]
        ]
        [Image Understanding, level2,
            [Change Detection \cite{dong2024changeclip, muhtar2024lhrsbot, ferrod2024towards}, leaf]
            [Object Detection \cite{wagner2020u, zhang2017separate, kampffmeyer2016semantic, garg2021isdnet, zang2024contextual, wang2024visionllm, zhao2024llm}, leaf]
            [Scene Description \cite{li2024cross, silva2024large, li2024learning, zhao2021high}, leaf]
        ]
    ]
    [Technical \\ Foundations, level1,
        [Remote Sensing Data \\ Characteristics, level2,
            [Temporal Aspects \cite{dong2024changeclip, chen2023self, zheng2024changen2, ravirathinam2024towards}, leaf]
            [Spectral Information \cite{roy2023multimodal, kieu2024multimodal, he2023foundation, xiao2024foundation}, leaf]
            [Spatial Resolution \cite{guo2024skysense, zhu2024cross, yang2024transcending, zhang2024segclip}, leaf]
        ]
        [Multi-Modal \\ Language Models, level2,
            [Integration Approaches \cite{liu2024bidirectional, sun2024cross, ma2022crossmodal, xu2023exploring}, leaf]
            [Learning Mechanisms \cite{guo2024skysense, zhu2024cross, zavras2024mind, wu2023self}, leaf]
            [Architecture Components \cite{silva2022remote, hoffmann2022multi, zhang2023fusion}, leaf]
        ]
    ]
]
\end{forest}
    \caption{Taxonomy of Multi-Modal Language Models for Remote Sensing}
    \label{fig:taxonomy}
\end{figure*}

\textbf{Fig \ref{fig:taxonomy}} presents a taxonomy of MLLM in remote sensing, categorizing the key components and their interrelationships as discussed in this review. The taxonomy is structured into six primary branches: Technical Foundations, Current Applications and Implementations, Datasets and Resources, Challenges and Limitations, and Future Directions and Conclusions. It serves as a guide for readers through the complexities of integrating and applying MLLMs to remote sensing tasks.

\section{Technical Foundations}
\subsection{Multi-Modal Language Models}
\subsubsection{Architecture Components}

MLLMs in remote sensing typically employ a dual-encoder architecture, integrating specialized components for processing both visual and textual information. These models often utilize a Transformer-based structure, which has proven effective in handling the complexities of remote sensing data \cite{silva2022remote}.

The visual component of these architectures commonly incorporates a Vision Transformer (ViT) or a convolutional neural network as the image encoder. This encoder is responsible for extracting relevant features from satellite imagery and other remote sensing data sources. The ViT architecture, in particular, has shown promise in adapting its attention mechanism and Transformer Encoder for remote sensing applications, allowing for efficient processing of spatial information \cite{hoffmann2022multi}.

On the language side, these models frequently employ BERT or similar Transformer-based language models as the text encoder. This component is crucial for understanding and processing textual queries, captions, or metadata associated with remote sensing imagery. The language encoder's self-attention mechanisms enable it to capture complex linguistic relationships and context \cite{zhang2023fusion}.

A key aspect of MLLMs in remote sensing is the integration of these visual and textual components. This integration is often achieved through cross-modal fusion techniques and sophisticated attention mechanisms. For instance, some architectures use self-attention multi-modal encoders that combine convolutional and recurrent elements with cross-modal fusion components \cite{silva2022remote}. Others employ unified Transformer models with cross-modal mixture experts, which allow for more nuanced interactions between visual and textual features \cite{liu2023unified}.

The attention mechanism plays a crucial role in these architectures, enabling the model to focus on relevant parts of both the image and text inputs. This is particularly important in remote sensing applications, where the ability to identify and focus on specific geographical features or phenomena is essential. The self-attention layers in these models allow for dynamic weighting of different input elements, enhancing the model's capacity to understand complex spatial and semantic relationships \cite{zhang2023fusion}.

\subsubsection{Learning Mechanisms}

Self-supervised learning techniques form the foundation of modern multi-modal models in remote sensing, enabling them to leverage vast amounts of unlabeled satellite imagery and associated text descriptions. These techniques allow models to learn meaningful representations without relying on expensive and time-consuming manual annotations. For instance, the SkySense model employs a Multi-Granularity Contrastive Learning approach for self-supervised pre-training on a large-scale multi-modal remote sensing imagery dataset, demonstrating the effectiveness of this learning mechanism in capturing complex spatiotemporal relationships \cite{guo2024skysense}.

Contrastive learning has emerged as a powerful paradigm in developing robust multi-modal representations for remote sensing applications. This approach helps establish strong relationships between visual and textual representations, enhancing the model's ability to align information across modalities. For example, Zhu et al. propose a cross-modal contrastive learning method that incorporates spatio-temporal context for multi-scale remote sensing image retrieval. Their approach uses a self-supervised contrastive loss to improve the model's capacity to capture correlations between different modalities and scales \cite{zhu2024cross}.

Cross-modal training is crucial in bridging the gap between different data modalities in remote sensing, such as optical imagery, synthetic aperture radar (SAR) data, and textual information. Zavras et al. address this challenge by developing a methodology for aligning distinct remote sensing imagery modalities with visual and textual modalities of CLIP (Contrastive Language-Image Pre-training). Their two-stage procedure involves robust fine-tuning of CLIP to handle distribution shifts, followed by cross-modal alignment of a remote sensing modality encoder \cite{zavras2024mind}.

The integration of intra-modal and cross-modal learning strategies has shown promise in improving the overall performance of multi-modal models. Wu et al. demonstrate this by combining self-supervised intra-modal and cross-modal contrastive learning for point cloud understanding, which is applicable to remote sensing data. Their approach leverages the complementary nature of different modalities to enhance feature representations and improve downstream task performance \cite{wu2023self}.

These learning mechanisms collectively enable MLLMs to capture the complex relationships between various data types in remote sensing. By leveraging self-supervised learning, contrastive approaches, and cross-modal training, these models can effectively process and interpret diverse Earth observation data, leading to more accurate and comprehensive analysis of remote sensing imagery.

\subsubsection{Integration Approaches}
Cross-modal feature fusion techniques play a crucial role in combining visual and textual representations in MLLMs for remote sensing applications. These techniques often employ carefully designed attention mechanisms and alignment strategies to ensure effective information exchange between modalities while preserving the unique characteristics of each data type. For instance, Liu and Wang propose a bidirectional feature fusion approach that incorporates enhanced alignment for multimodal semantic segmentation of remote sensing images. Their method uses cross-attention mechanisms to maintain the integrity of single-modal information while achieving effective feature alignment across modalities \cite{liu2024bidirectional}.

Modal alignment is a critical aspect of integration approaches, as it ensures that features from different modalities are properly synchronized and correlated. Sun et al. introduce a cross-modal pre-aligned method that incorporates both global and local information for remote sensing image and text retrieval. Their approach employs a multi-modal encoder for cross-modal feature fusion and utilizes cross-attention mechanisms to fully exploit the relationships between different modalities. This pre-alignment strategy helps to establish a strong foundation for subsequent feature integration and retrieval tasks \cite{sun2024cross}.

Cross-attention mechanisms have emerged as a powerful tool for facilitating information exchange and alignment between different modalities in remote sensing applications. Ma et al. present a crossmodal multiscale fusion network for semantic segmentation of remote sensing data, which leverages cross-attention and cross-transformer techniques for multimodal data fusion. Their approach addresses challenges in feature extraction, alignment, and fusion across different modalities, demonstrating the effectiveness of cross-attention in capturing complex inter-modal relationships \cite{ma2022crossmodal}.

To further enhance the integration of multi-modal remote sensing data, researchers have explored self-supervised learning techniques combined with asymmetric attention fusion. Xu et al. propose a method that utilizes cross-attention mechanisms to transfer features between modalities during pre-training. This approach addresses the challenge of integrating multi-modal data with strong inter-modal complementarity, allowing the model to learn more robust and generalizable representations \cite{xu2023exploring}.

These integration approaches collectively contribute to the development of more sophisticated and effective MLLMs for remote sensing applications. By leveraging advanced feature fusion techniques, modal alignment strategies, and cross-attention mechanisms, these models can better capture the complex relationships between visual and textual information in Earth observation data, leading to improved performance across various remote sensing tasks.

\subsection{Remote Sensing Data Characteristics}
\subsubsection{Spatial Resolution Considerations}
Spatial resolution in remote sensing imagery varies significantly, ranging from sub-meter to kilometer-scale ground sampling distances. This variability presents unique challenges for MLLMs, necessitating adaptive processing approaches to maintain model performance across different scales. The SkySense model, for instance, addresses this issue by employing a factorized multi-modal spatiotemporal encoder that learns representations across different modal and spatial granularities. This approach enables the model to effectively process and interpret remote sensing data at various spatial resolutions, demonstrating the importance of multi-scale analysis in handling diverse Earth observation imagery \cite{guo2024skysense}.

To tackle those challenges, researchers have developed multi-scale analysis techniques that allow models to adapt to different levels of detail in remote sensing imagery. Zhu et al. propose a cross-modal contrastive learning approach that incorporates spatio-temporal context for correlation-aware multi-scale remote sensing image retrieval. Their method addresses the spatial resolution variability by considering multiple scales during the learning process, enabling more robust and adaptive representations of remote sensing data \cite{zhu2024cross}.

The integration of multi-scale alignment methods has shown promise in improving the performance of MLLMs across different spatial resolutions. Yang et al. introduce a Multi-Scale Alignment (MSA) method for remote sensing image-text retrieval, which includes a Multi-scale Cross-Modal Alignment Transformer (MSCMAT). This approach computes cross-attention between single-scale image features and localized text features while integrating global textual context. By considering multiple scales in the alignment process, the model can better handle the diverse spatial resolutions encountered in remote sensing applications \cite{yang2024transcending}.

For high-resolution remote sensing imagery, adaptive processing techniques are particularly crucial. Zhang et al. present SegCLIP, a multimodal approach for high-resolution remote sensing semantic segmentation that incorporates a multi-scale prototype transformer decoder. This method specifically addresses the challenges associated with processing high-resolution remote sensing imagery by employing multi-scale analysis techniques. By adapting to different levels of detail within the high-resolution data, SegCLIP demonstrates the effectiveness of multi-scale approaches in maintaining model performance across varying spatial resolutions \cite{zhang2024segclip}.

These adaptive processing approaches and multi-scale analysis techniques collectively enable MLLMs to effectively handle the wide range of spatial resolutions encountered in remote sensing applications. By incorporating methods that can adapt to different scales and levels of detail, these models can maintain their performance and accuracy across diverse remote sensing datasets, ultimately improving their utility in Earth observation and analysis tasks.

\subsubsection{Spectral Information}
Remote sensing systems capture data across multiple spectral bands, ranging from visible light to infrared wavelengths, providing rich information about Earth's surface properties. This spectral diversity is a key component in developing advanced MLLMs for remote sensing applications. Multispectral (MS) sensors measure reflected light within specific wavelength ranges, offering a comprehensive view of the Earth's surface that goes beyond what is visible to the human eye \cite{roy2023multimodal}. These sensors typically capture data in several discrete bands, each sensitive to a particular portion of the electromagnetic spectrum.

Hyperspectral sensors, on the other hand, collect data across a much larger number of narrower spectral bands, often numbering in the hundreds. This high spectral resolution allows for more detailed analysis of surface properties and materials. The integration of hyperspectral data into MLLMs presents both opportunities and challenges, as it provides a wealth of information but also increases computational complexity \cite{kieu2024multimodal}.

The selection and processing of spectral bands play a crucial role in extracting meaningful features for multi-modal analysis. Band selection techniques are employed to identify the most informative spectral bands for specific applications, reducing data dimensionality while preserving essential information. This process is particularly important when dealing with hyperspectral data, where the high number of bands can lead to the "curse of dimensionality" and increased processing time \cite{he2023foundation}.

Multi-modal fusion approaches, such as those employing transformer architectures, have shown promise in effectively combining spectral information with other data modalities like synthetic aperture radar (SAR) or textual data. These fusion techniques allow models to leverage the complementary nature of different spectral ranges and data types, leading to improved performance in tasks such as image classification, object detection, and semantic segmentation \cite{xiao2024foundation}.

The development of foundation models for remote sensing has further advanced the utilization of spectral information. These models, inspired by large language models, aim to learn generalizable representations from diverse spectral data sources. By incorporating varying spatial and spectral resolutions, foundation models can potentially address the challenges posed by the complexity of Earth's environments and the distinctive feature patterns found in remote sensing data \cite{xiao2024foundation}.

As MLLMs for remote sensing continue to evolve, the effective integration and interpretation of spectral information remain key areas of research. Future advancements may focus on developing more sophisticated band selection algorithms, improving the fusion of multispectral and hyperspectral data with other modalities, and enhancing the ability of models to handle the temporal dynamics inherent in Earth observation data.

\subsubsection{Temporal Aspects}
MLLMs for remote sensing are increasingly incorporating temporal aspects to enhance their ability to analyze and interpret data over time. These models track dynamic processes and detect changes in the Earth's surface, addressing challenges such as irregular sampling intervals and seasonal variations.

One significant application of temporal aspects in multi-modal models is change detection. ChangeCLIP, for instance, utilizes multimodal vision-language representation learning to identify changes in satellite imagery over time \cite{dong2024changeclip}. This approach demonstrates the potential of integrating temporal information with both visual and textual data to improve the accuracy and interpretability of change detection tasks in remote sensing.

The analysis of satellite image time series presents unique challenges due to the irregular nature of data acquisition. To address this, researchers have developed self-supervised learning techniques that can handle the complexities of temporal data in remote sensing \cite{chen2023self}. These methods aim to extract meaningful temporal patterns and changes from satellite imagery, even when dealing with non-uniform sampling intervals or missing data points.

Advancements in generative models have also contributed to the field of temporal remote sensing analysis. Changen2, a multi-temporal remote sensing generative change foundation model, works with time series of remote sensing images to generate globally distributed change detection datasets \cite{zheng2024changen2}. This approach not only aids in change detection but also has the potential to augment existing datasets, addressing the scarcity of labeled temporal data in remote sensing.

Integrating knowledge-guided approaches with multimodal foundation models pushes the boundaries of spatio-temporal remote sensing. These models can handle a variety of tasks that require temporal understanding, such as land-use land cover change detection and crop yield prediction \cite{ravirathinam2024towards}. By incorporating domain knowledge and temporal information, these models can provide more accurate and contextually relevant interpretations of remote sensing data over time.

\section{Current Applications and Implementations}
\subsection{Image Understanding Tasks}
\subsubsection{Scene Description}
Multi-modal models for remote sensing have made significant advancements in generating detailed natural language descriptions of satellite imagery scenes. These models leverage sophisticated techniques to capture spatial relationships and key geographic features, enabling automated interpretation of complex Earth observation data.

Recent research has focused on improving the semantic understanding and caption quality of remote sensing images. Li et al. (2024) proposed a two-stage approach that combines cross-modal retrieval with semantic refinement \cite{li2024cross}. This method addresses limitations in semantic understanding by extracting supplementary information from related remote sensing tasks, resulting in more accurate and contextually relevant image descriptions.

The integration of large language models (LLMs) with remote sensing image captioning has shown promising results. Silva et al. (2024) explored the use of LLMs for both image captioning and text-image retrieval tasks \cite{silva2024large}. Their approach incorporates semantic segmentation techniques to enhance the model's ability to identify and describe distinct regions within satellite imagery. This integration allows for a more comprehensive understanding of the scene, capturing both visual and semantic information.

To further improve captioning accuracy, researchers have developed consensus-aware semantic knowledge approaches. Li et al. (2024) introduced a method that utilizes multi-modal representations based on semantic-level relational features and visual-semantic contextual vectors \cite{li2024learning}. This approach enables the model to generate more coherent and contextually appropriate captions by considering the consensus among different semantic aspects of the image.

Structured attention mechanisms have also been applied to high-resolution remote sensing image captioning. Zhao et al. (2021) developed a model that incorporates semantic segmentation techniques within a structured attention framework \cite{zhao2021high}. This approach allows for more precise identification of distinct regions and their characteristics, leading to improved captioning performance for complex, high-resolution satellite imagery.

\subsubsection{Object Detection}

Object detection and instance segmentation play a pivotal role in remote sensing for urban monitoring and environmental assessment by enabling the identification and delineation of entities within satellite and aerial imagery, including buildings, vehicles, and natural features. Techniques such as U-net-id have been effective in extracting detailed building information for urban planning and monitoring, aiding in assessing land use and infrastructure development \cite{wagner2020u}. Additionally, segmenting multi-temporal high-resolution imagery facilitates object-based change detection for tracking urban expansion and unauthorized construction \cite{zhang2017separate}. Deep CNNs address challenges in detecting smaller objects while maintaining high accuracy, with advancements allowing pixel-scale uncertainty quantification \cite{kampffmeyer2016semantic}. Further enhancements from models like ISDNet contribute to detailed scene analysis for applications in traffic monitoring and green space assessment \cite{garg2021isdnet}.

Recent advancements in MLLMs have further improved object detection. For instance, ContextDET employs a generate-then-detect framework that combines a visual encoder, a pre-trained language model, and a visual decoder to identify and locate objects within human vocabulary, demonstrating improved performance in contextual object detection tasks \cite{zang2024contextual}. Similarly, VisionLLM utilizes a language-guided image tokenizer and an LLM-based decoder to seamlessly integrate various vision-centric tasks, including object detection and instance segmentation, into a unified model \cite{wang2024visionllm}. Additionally, LLM-Optic leverages large language models to enhance visual grounding by interpreting complex text queries and accurately identifying corresponding objects in images, achieving state-of-the-art zero-shot visual grounding capabilities \cite{zhao2024llm}. These developments underscore the potential of MLLMs to advance object detection by effectively combining visual and textual information.

\subsubsection{Change Detection}

MLLMs enhance change detection in remote sensing by integrating visual and textual data. ChangeCLIP uses vision-language representation learning to identify surface changes from bitemporal images, achieving state-of-the-art performance on multiple datasets \cite{dong2024changeclip}. Similarly, LHRS-Bot utilizes volunteered geographic information and remote sensing images to perform various tasks, including change detection, demonstrating the potential of MLLMs in this domain \cite{muhtar2024lhrsbot}. Additionally, the development of a multimodal framework for remote sensing image change retrieval and captioning has enabled continuous monitoring of landscape changes, facilitating environmental protection and disaster monitoring \cite{ferrod2024towards}. These advancements underscore the efficacy of MLLMs in analyzing temporal sequences of satellite imagery to identify and characterize surface modifications over time.

\subsection{Cross-Modal Applications}
\subsubsection{Text-to-Image Retrieval}
Text-to-image retrieval in remote sensing enables users to search satellite imagery databases using natural language queries. This cross-modal search aligns textual descriptions with visual features, facilitating efficient access to relevant Earth observation data. Recent advancements have focused on developing models that bridge the semantic gap between text and images. For instance, Yuan et al. proposed a method that integrates global and local information to enhance retrieval accuracy, addressing the challenge of capturing both overall scene context and specific object details \cite{yuan2022remote}.

To improve retrieval performance, some approaches incorporate external knowledge. Mi et al. introduced a knowledge-aware text-image retrieval method that enriches textual queries with information from external knowledge graphs, thereby reducing information asymmetry between modalities \cite{mi2024knowledge}. Additionally, Pan et al. presented a prior instruction representation learning paradigm, leveraging prior knowledge to instruct adaptive learning of vision and text representations, which enhances retrieval effectiveness \cite{pan2024pir}.

Efforts have also been made to optimize retrieval efficiency. Mikriukov et al. developed a deep unsupervised contrastive hashing method for large-scale cross-modal text-image retrieval, enabling fast and memory-efficient searches without requiring labeled training samples \cite{mikriukov2022deep}. These advancements collectively contribute to more accurate and efficient text-to-image retrieval systems in remote sensing, facilitating better utilization of satellite imagery for various applications.

\subsubsection{Image-to-Text Generation}
Automated caption generation for satellite imagery translates visual data into descriptive text, enhancing the accessibility and interpretability of remote sensing information. Recent advancements have employed transformer-based architectures to improve the quality of generated captions. Shen et al. introduced a model that integrates transformers with reinforcement learning, effectively capturing semantic information and addressing overfitting issues associated with limited datasets \cite{shen2020remote}.

Incorporating attention mechanisms has further refined image-to-text generation. Li et al. proposed a multi-level attention model that mimics human cognitive processes by focusing on different image regions and corresponding textual elements, resulting in more accurate and contextually relevant descriptions \cite{li2020multi}. Additionally, Silva et al. explored the application of large language models for captioning and retrieving remote sensing images, demonstrating that pre-trained models can be adapted to generate detailed and coherent captions for satellite imagery \cite{silva2024large}.

Visual grounding techniques play a crucial role in ensuring that generated captions accurately reflect the content and spatial relationships within imagery. Zhan et al. introduced the task of visual grounding for remote sensing data, aiming to localize referred objects in images based on natural language descriptions. Their work includes the development of a large-scale benchmark dataset and a transformer-based model that leverages multi-level cross-modal feature learning to enhance grounding accuracy \cite{zhan2023rsvg}. These advancements collectively contribute to more precise and informative caption generation systems for satellite imagery.

\subsubsection{Visual Question Answering}
Visual Question Answering (VQA) systems in remote sensing enable users to query satellite imagery using natural language, facilitating interactive analysis of Earth observation data. These systems integrate visual and textual information to provide contextually relevant responses. Lobry et al. introduced the RSVQA dataset, which pairs remote sensing images with corresponding questions and answers, serving as a benchmark for developing and evaluating VQA models in this domain \cite{lobry2020rsvqa}.

To enhance the performance of VQA systems, researchers have explored various model architectures. Zheng et al. proposed the Mutual Attention Inception Network (MAIN), which employs mutual attention mechanisms to effectively capture interactions between image features and textual queries, improving answer accuracy \cite{zheng2021mutual}. Similarly, Siebert et al. developed a Multi-Modal Fusion Transformer that leverages transformer-based architectures to fuse visual and textual modalities, demonstrating improved performance on VQA tasks \cite{siebert2022multi}.

Recent advancements have focused on lightweight models to reduce computational requirements. Hackel et al. introduced LiT-4-RSVQA, a transformer-based architecture designed for efficient VQA in remote sensing. This model achieves competitive accuracy while significantly reducing computational demands, making it suitable for operational scenarios \cite{hackel2023lit}. These developments collectively contribute to more effective and accessible VQA systems for remote sensing applications.

\section{Datasets and Resources}
\subsection{Benchmark Datasets}
Benchmark datasets are essential for developing and evaluating MLLMs in remote sensing. The RSICap dataset, introduced by Hu et al., comprises 2,585 high-resolution remote sensing images, each annotated with five detailed captions. This dataset facilitates tasks such as image captioning and visual question answering by providing rich semantic information \cite{hu2023rsgpt}.

To address the need for larger datasets, Zhang et al. developed RS5M, a collection of 5 million remote sensing images paired with English descriptions. This dataset was created by filtering and captioning existing datasets, enabling the fine-tuning of vision-language models for tasks like zero-shot classification and cross-modal retrieval \cite{zhang2023rs5m}.

Another significant contribution is ChatEarthNet, presented by Yuan et al., which offers 163,488 image-text pairs with captions generated by ChatGPT-3.5, along with an additional 10,000 pairs from ChatGPT-4V. This dataset provides high-quality natural language descriptions for global-scale satellite data, enhancing the training and evaluation of vision-language models in remote sensing \cite{yuan2024chatearthnet}

\subsection{Training Resources}

The development of remote sensing MLLMs has been significantly advanced by the availability of pre-trained models and specialized training frameworks. These resources reduce computational demands and expedite the creation of new applications. SkySense is a multi-modal remote sensing foundation model pre-trained on a large-scale dataset comprising 21.5 million temporal sequences from high-spatial-resolution optical images, medium-resolution temporal multispectral imagery, and temporal synthetic aperture radar imagery \cite{guo2024skysense}. SkySense's architecture includes a factorized multi-modal spatiotemporal encoder, enabling it to handle diverse remote sensing tasks effectively.

In addition to pre-trained models, specialized training frameworks have been developed to support multi-modal learning in remote sensing. The Language-Guided Visual Prompt Compensation framework addresses modality absence by integrating language guidance into visual prompt learning, enhancing the model's robustness in scenarios where certain modalities are missing \cite{huang2024language}. This approach utilizes the complementary nature of language and visual data, improving the adaptability of multi-modal systems.

Furthermore, self-supervised learning techniques have been employed to pre-train models on multi-modal remote sensing data, reducing the reliance on labeled datasets. Berg et al. proposed a joint multi-modal self-supervised pre-training method that learns representations from unlabelled data, facilitating downstream tasks such as methane source classification \cite{berg2023joint}. This method demonstrates the potential of self-supervised learning in extracting meaningful features from multi-modal remote sensing data, thereby enhancing the efficiency and effectiveness of model training.

\subsection{Evaluation Metrics}
Evaluating the performance of MLLMs, particularly in generating textual descriptions from satellite imagery, relies on several established metrics. The BLEU (Bilingual Evaluation Understudy) score assesses the precision of n-grams in the generated text against reference descriptions, providing insight into the model's ability to produce accurate and fluent language \cite{papineni2002bleu}. Similarly, the METEOR (Metric for Evaluation of Translation with Explicit ORdering) metric evaluates both precision and recall, incorporating synonymy and stemming to better align with human judgment \cite{banerjee2005meteor}.

The CIDEr (Consensus-based Image Description Evaluation) metric is tailored for image captioning tasks. It measures the similarity of a generated description to a set of reference captions by considering term frequency-inverse document frequency (TF-IDF) weighting, thereby emphasizing the importance of informative words \cite{vedantam2015cider}. This approach ensures that the generated text aligns closely with human descriptions, capturing the nuances of the visual content.

These metrics have been applied to assess the quality of descriptions generated from satellite images. For instance, the EM-VLM4AD model, designed for visual question answering in autonomous driving, was evaluated using BLEU-4, METEOR, and CIDEr scores, demonstrating the applicability of these metrics in multi-modal tasks \cite{gopalkrishnan2023em}. Such evaluations are crucial for developing models that effectively bridge the gap between visual data and natural language, enhancing the interpretability and usability of satellite imagery.




\section{Challenges and Limitations}

\subsection{Technical Challenges}

\subsubsection{Computational Requirements}
Processing high-resolution satellite imagery with MLLMs requires substantial computational resources. The integration of diverse data modalities, such as text and images, increases the complexity of these models, leading to higher computational demands. For instance, the training of large-scale MLLMs like LLaVA and multi-modal Phi-2 necessitates significant GPU resources to handle the extensive data and model parameters involved \cite{basu2024understanding}.

Memory constraints further limit the deployment of MLLMs in practical applications. High-resolution satellite images contain vast amounts of data, and processing them alongside textual information requires models with large memory capacities. To address this issue, researchers have explored parameter-efficient fine-tuning (PEFT) techniques. Context-PEFT, for example, introduces a novel framework that reduces the number of trainable parameters and GPU memory consumption without compromising performance \cite{hadji2023context}.

Efforts to optimize MLLMs for resource-constrained environments have also been made. EM-VLM4AD is an efficient, lightweight, multi-frame vision-language model designed for autonomous driving applications. This model requires significantly less memory and computational power while achieving higher performance metrics compared to previous approaches, demonstrating the potential for deploying MLLMs in resource-limited settings \cite{gopalkrishnan2024multi}. These advancements highlight the ongoing efforts to balance computational requirements and model performance in the development of MLLMs for remote sensing applications.

\subsubsection{Model Scalability}
Scaling MLLMs to handle increasing data volumes and higher resolutions presents significant technical challenges. The relationship between model size and performance is complex, influencing practical deployment decisions. Recent studies have explored this relationship, revealing that while larger models often achieve better performance, they also require more computational resources. For instance, Aghajanyan et al. investigated scaling laws for generative mixed-modal language models, demonstrating that performance improvements diminish as model size increases, highlighting the need for efficient scaling strategies \cite{aghajanyan2023scaling}.

To address these challenges, researchers have proposed various approaches to enhance model scalability. One such method is multi-resolution training, which allows models to process inputs at different resolutions, thereby improving efficiency without compromising performance. Tian et al. introduced ResFormer, a model that employs multi-resolution training to scale vision transformers effectively, enabling them to handle high-resolution inputs with reduced computational costs \cite{tian2022resformer}. Additionally, integrating multi-modal data poses unique scalability issues, as models must effectively process and align information from diverse sources. Aghajanyan et al. also examined scaling laws for generative mixed-modal language models, providing insights into how model size and data complexity interact in multi-modal settings \cite{aghajanyan2023scaling}.

Efforts to improve scalability have also focused on optimizing model architectures and training strategies. For example, Liu et al. developed SPHINX-X, a family of multi-modal large language models that scale data and parameters efficiently. Their approach involves modifying the model architecture to remove redundant components and simplifying the training process, resulting in improved scalability and performance \cite{liu2024sphinx}. These advancements underscore the importance of developing scalable MLLMs capable of handling large-scale, high-resolution data in practical applications.

\subsubsection{Data Quality Issues}
Satellite imagery frequently contains noise and artifacts due to atmospheric disturbances, sensor limitations, and environmental factors, which can adversely affect the performance MLLMs in remote sensing applications. These imperfections introduce inconsistencies that challenge the models' ability to accurately interpret and analyze the data. For instance, Pasquarella et al. developed a comprehensive quality assessment framework for optical satellite imagery, utilizing weakly supervised learning to identify and quantify various image artifacts \cite{pasquarella2023comprehensive}.

To mitigate the impact of noise, advanced denoising techniques have been proposed. Thai et al. introduced the Riesz-Quincunx-UNet Variational Auto-Encoder, which effectively reduces noise in satellite images by integrating wavelet transforms with deep learning architectures \cite{thai2022riesz}. Similarly, Song et al. presented an unsupervised denoising method using a Wavelet Subband CycleGAN, demonstrating significant improvements in image quality without requiring paired training data \cite{song2020unsupervised}.

Quality assessment procedures are essential to address variations in atmospheric conditions and sensor characteristics. Mikriukov et al. proposed a deep unsupervised contrastive hashing method for large-scale cross-modal text-image retrieval in remote sensing, which includes mechanisms to assess and enhance data quality \cite{mikriukov2022deep}. These approaches underscore the importance of robust data quality assessment and enhancement techniques to ensure the reliability and accuracy of MLLMs in processing satellite imagery.

\subsection{Application-Specific Challenges}

Domain adaptation remains a significant challenge in remote sensing, as models trained on specific geographic regions or sensor types often struggle when applied to new domains. To address this, techniques like GeoRSCLIP, which fine-tunes vision-language models on remote sensing datasets such as RS5M, have been introduced to bridge the gap between general and domain-specific tasks \cite{zhang2023rs5m}. Incorporating external knowledge, such as through knowledge-aware retrieval methods that integrate information from external graphs, further enhances adaptation by mitigating information asymmetry \cite{mi2024knowledge}. Prior instruction representation learning has also been employed to guide adaptive learning for improved cross-modal retrieval \cite{pan2024pir}. Additionally, methods like unsupervised contrastive hashing enable efficient, large-scale text-image retrieval, supporting more effective use of satellite data without extensive labeled samples \cite{mikriukov2022deep}.

\section{Future Directions and Conclusions}
\subsection{Research Opportunities and Technological Advancements}

Advancements in MLLMs for remote sensing present significant research opportunities. Integrating multi-temporal and multi-scale data is essential for analyzing dynamic environments. Recent studies like SkySense have explored the fusion of diverse data sources to enhance model performance \cite{guo2024skysense}. Similarly, EarthGPT is a universal MLLM designed for multi-sensor image comprehension, effectively handling data from different remote sensing modalities \cite{zhang2024earthgpt}. Additionally, the adoption of hardware accelerators, such as GPUs and TPUs, has been shown to significantly speed up model training and inference processes, facilitating the deployment of MLLMs in real-world scenarios \cite{guo2024skysense}. Continued research in these areas is essential for broadening the applications and accessibility of MLLMs in remote sensing.

\subsection{Potential Applications and Recommendations}
Emerging applications of MLLMs are reshaping remote sensing by enabling real-time environmental monitoring, facilitating early detection of deforestation and pollution. In disaster response, MLLMs enhance situational awareness through rapid damage assessment from diverse data sources \cite{guo2024skysense}. Urban planning benefits through detailed land-use mapping and infrastructure analysis, promoting sustainable development \cite{zhang2024earthgpt}.

Adopting MLLMs in industry requires developing models tailored for specific use cases, improving precision and context-awareness in sectors like agriculture, forestry, and maritime surveillance \cite{guo2024skysense, yuan2024surveillance}. For instance, models focused on agricultural monitoring can analyze crop health and predict yields, while those specialized for maritime surveillance can detect illegal fishing \cite{zhang2024earthgpt, zheng2023marinegpt}.

Future research shall prioritize model efficiency and domain adaptation to generalize across different remote sensing datasets \cite{guo2024skysense, yuan2024surveillance}. Optimizing computational resources and establishing standardized evaluation methods will drive consistent progress and enhance practical applicability while improve overall safety \cite{zhang2024earthgpt, peng2024securing, peng2024jailbreaking}. Collaboration between academia and industry is crucial for developing robust models that meet real-world requirements.

\bibliographystyle{ieeetr}  
\bibliography{references} 


\end{document}